\documentclass[a4paper, 10pt, conference]{ieeeconf}      

\usepackage{times}
\usepackage{epsfig}
\usepackage{graphicx}
\usepackage{amsmath}
\usepackage{amssymb}
\usepackage{textcomp}
\usepackage{multirow}
\usepackage{adjustbox}

\usepackage{multirow}
\usepackage{colortbl}
\usepackage{hhline}
\usepackage{subcaption}
\usepackage{verbatim}

\usepackage{gensymb}
\usepackage{xcolor}
\usepackage{hyperref}

\IEEEoverridecommandlockouts                              

\title{\LARGE \bf
Spatio-Temporal Multi-Task Learning Transformer for Joint Moving Object Detection and Segmentation}

\author{Eslam Mohamed$^{1}$, Ahmad El Sallab$^{1}$ \\ %
$^{1}$Valeo R\&D Cairo, Egypt \\
         {\tt \small\{eslam.mohamed-abdelrahman, ahmad.el-sallab\}@valeo.com}
}

\begin{document}

\maketitle
\thispagestyle{empty}
\pagestyle{empty}

\begin{abstract}
Moving objects have special importance for Autonomous Driving tasks. Detecting moving objects can be posed as Moving Object Segmentation, by segmenting the object pixels, or Moving Object Detection, by generating a bounding box for the moving targets. In this paper, we present a Multi-Task Learning architecture, based on Transformers, to jointly perform both tasks through one network. Due to the importance of the motion features to the task, the whole setup is based on a Spatio-Temporal aggregation. We evaluate the performance of the individual tasks architecture versus the MTL setup, both with early shared encoders, and late shared encoder-decoder transformers. For the latter, we present a novel joint tasks query decoder transformer, that enables us to have tasks dedicated heads out of the shared model. To evaluate our approach, we use the KITTI  MOD \cite{siam2018modnet} data set. Results show 1.5\% mAP  improvement for Moving Object Detection, and 2\% IoU improvement for Moving Object Segmentation, over the individual tasks networks.  
\end{abstract}

\section{INTRODUCTION}
\begin{figure*}[ht!]
\begin{center}
\centerline{\includegraphics[width=170mm]{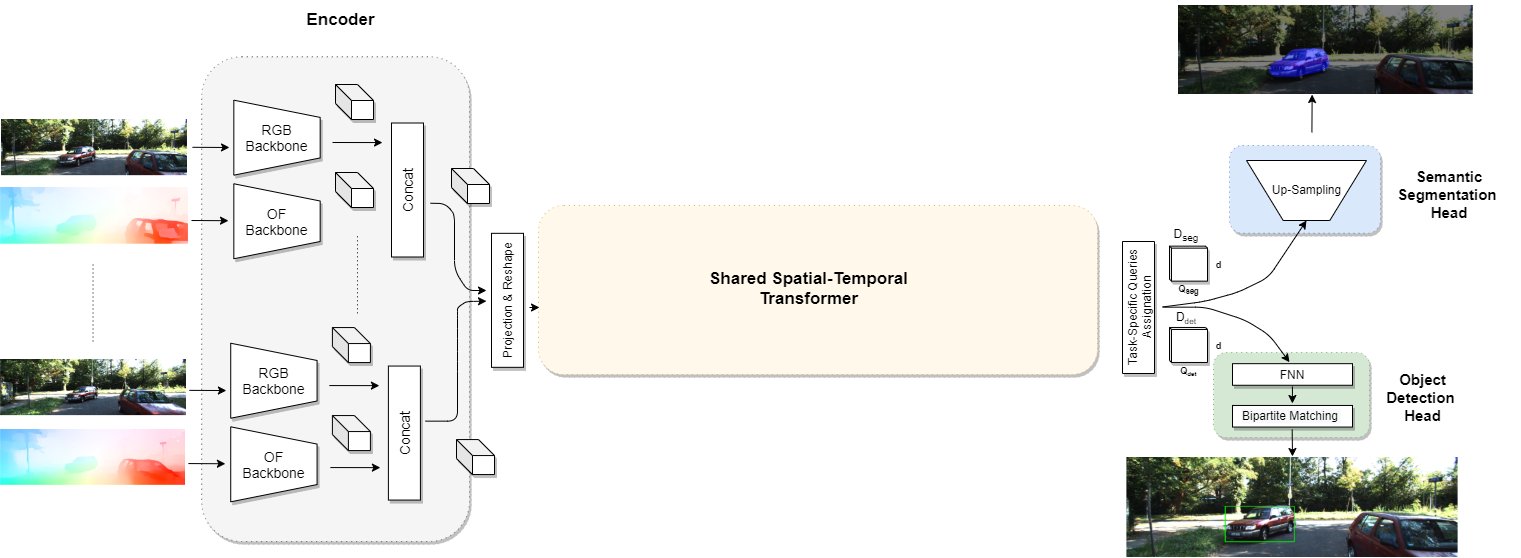}}
\caption{Spatio-Temporal Multi-Task Learning Transformer Model Architecture for Joint MOD + MOSeg}
\label{mtl_arch}
\end{center}
\end{figure*}

Moving object Detection (MOD) or Segmentation (MOSeg), are crucial tasks for autonomous  vehicles as they can be used to segment objects depending on their motion features. The input is a video sequence of frames. Hence, it is important to take advantage of the temporal correlations between input features. Our goal is to merge both tasks in one Multi-Task Learning (MTL) architecture, learning both jointly.

For many years, ConvNets have been the architecture of choice in computer vision in general, and for performing object detection tasks in particular. Recently, transformers have shown good results compared to ConvNets, in object detection, \cite{carion2020end}. The attention-is-all you need transformers \cite{vaswani2017attention}, introduced in the NLP domain, is a natural fit to handle sequential input, which is the sequence of words in the case of NLP. Transformers have been recently used for object detection, like DETR \cite{carion2020end}, and segmentation as in \cite{xie2021segmenting} \cite{zheng2020rethinking}. 

In both setups, the input image is first encoded in a traditional ConvNet backbone and then passed through a Transformer encoder, which performs self-attention across the spatial dimension, followed by a specialized Transformer decoder. In case of detection, like in DETR, the  decoder is based on learnable object queries attention, followed by Hungarian matching and bi-partite loss \cite{carion2020end}. While for segmentation, the decoder is based on querying the segmented class labels as in \cite{xie2021segmenting}, followed by an upsampling ConvNet based head. In our MTL setup shown in Figure \ref{mtl_arch}, we aim to have shared transformers, followed by specialized heads. We evaluate two setups, the first is to have a shared encoder, followed by specialized decoders and the corresponding output heads. In the second setup, we have a shared encoder and decoder transformers, followed by specialized output heads. In the latter setup, we propose a novel joint tasks query attention, which concatenates both tasks queries; objects, and segmented classes, in one query vector, which is later split in the corresponding task heads.

To handle the temporal dimension, we perform a Spatio-temporal sequence-to-sequence mapping. We perform early temporal aggregation of the spatial features, resulting in a temporal trace of features at each spatial location.

We evaluate the MTL model on both the individual tasks, MOSeg and MOD, and their joint setup. We use the published dataset KITTI MOD \cite{siam2018modnet} which includes the motion masks. 
The rest of the paper is organized as follows, first, we discuss the related work, followed by the details of the proposed model, starting from the dissection of the vanilla DETR, to the needed modifications to transform it into ST-DETR, which leads to the discussion of the two architecture variants: early vs. late temporal aggregation. Then we present the experimental setup for the various experiments we conducted for every contribution, and finally, we conclude with the discussion of the main findings, insights, and outcomes.

\section{RELATED WORK} \label{sec:relatedwork}

\textbf{ConvNet based methods} For object detection, Convolutional based methods are classically divided into: 1) Two-stage detectors, based on separate Region Proposals, like R-CNN \cite{cai2019cascade}, Fast R-CNN \cite{girshick2015fast} and Faster R-CNN \cite{ren2015faster} and 2) One-stage or One-shot methods, which merges the regions proposals and refinement in one network, like SSD \cite{liu2016ssd}, and the popular YOLO architectures: \cite{redmon2016you, redmon2017yolo9000, redmon2018yolov3, bochkovskiy2020yolov4}. Both approaches require a tedious post-processing pipeline, during training to assign the ground truth to labels, and during inference to refine the predictions using Non-Maximal suppression (NMS). While for semantic segmentation, the most popular ConvNet is based on the encoder-decoder architecture, based on the idea of fully convolutional networks (FCN) \cite{long2015fully}, \cite{ronneberger2015u} and \cite{chen2017deeplab}.

\textbf{Transformer based methods}: For object detection in the DEtection TRansformer (DETR) \cite{carion2020end} the input image is treated as sequence of spatial features. This enables the extension of the traditional transformer, previously used in NLP \cite{vaswani2017attention}, in computer vision problems. Full attention mechanisms are employed to extract feature interactions in an end-to-end architecture, followed by bi-partite matching that enables the replacement of the complex post-processing pipeline in the corresponding ConvNet architectures during training. The ground truth to prediction matching is treated as an association problem and solved using the Hungarian algorithm, producing one-one mapping that can be used to calculate the loss. 
While Panoptic segmentation is possible directly in DETR \cite{carion2020end}, using object queries, semantic masks require different architecture changes. Recent works extend the encoder-decoder architecture using transformers. In SETR \cite{zheng2020rethinking}, the encoder is kept convolution based same as in FCN \cite{long2015fully}, while the decoder is based on the transformer decoder architecture, with the learnable queries using progressive upsampling. The same idea is used in TransUNET \cite{chen2021transunet}, following the UNet architecture with skip connections between the encoder and decoder. In \cite{xie2021segmenting}, a full transformer encoder-decoder architecture is used, which is the closest to the architecure used for MOSeg in this paper. However, in \cite{xie2021segmenting}, the decoded segmentation mask is taken as the decoder attention weights directly, while in our case, we keep the Multi-Head attention query-key-value structure to decode the final segmentation mask.

\textbf{Spatio-Temporal methods} Like ConvNets in spatial computer vision, Recurrent models have been the architecture of choice for sequence models, especially in NLP. In computer vision, ConvNets and LSTM mixed architectures, like ConvLSTM have been used to handle both the spatial and temporal nature of videos, like in Moving Object Detection (MOD) \cite{siam2017modnet,siam2018modnet}, and Instance Moving Object Segmentation \cite{mohamed2020instancemotseg} tasks. 
Recently full attention transformers \cite{vaswani2017attention} are replacing RNN, LSTM, and GRU in NLP, taking advantage of the parallel encoding process, which removes the sequential nature of recurrent models. This motivates our work here to extend the DETR to handle also the temporal dimension, and to replace the ConvLSTM models to take advantage of the fast nature of the transformer architectures.

\textbf{Multi-Task Learning} is an inductive learning approach that exploits shared information among different tasks to learn refined shared features that boost the learning process concerning the convergence time, model architecture size, and accuracy. Determining which parts from the model will be shared across different tasks, which tasks could be trained jointly, and the training recipe for multi-task learning are challenging aspects.
In \cite{zhang2014facial} \cite{dai2016instance} \cite{zhao2018modulation} \cite{liu2019end} \cite{ma2018modeling} a shared encoder is used while learning different tasks, that is followed by a task specific output head.
Instead of using a shared feature extractor only, encoder, with task-specific output heads, \cite{misra2016cross} \cite{ruder2019latent} follow a cross-talk technique, whereas a totally separate network is used for each task while comprising features from corresponding layers from each task to share their representation while the training process. Despite the cross-talk technique may outputs a better feature representation than sharing the feature extraction part only however more parameters are needed in this approach which doesn't fit in various real-world applications where time and memory constraints will not be satisfied.
Mti-net \cite{vandenhende2020mti} and Pad-net \cite{xu2018pad} follow another MTL technique where each task takes advantage of each other by producing initial predictions for each task using a separate model for each one, then consolidating these predictions while producing the refined outputs.

\begin{figure*}
\begin{center}
\centerline{\includegraphics[width=170mm]{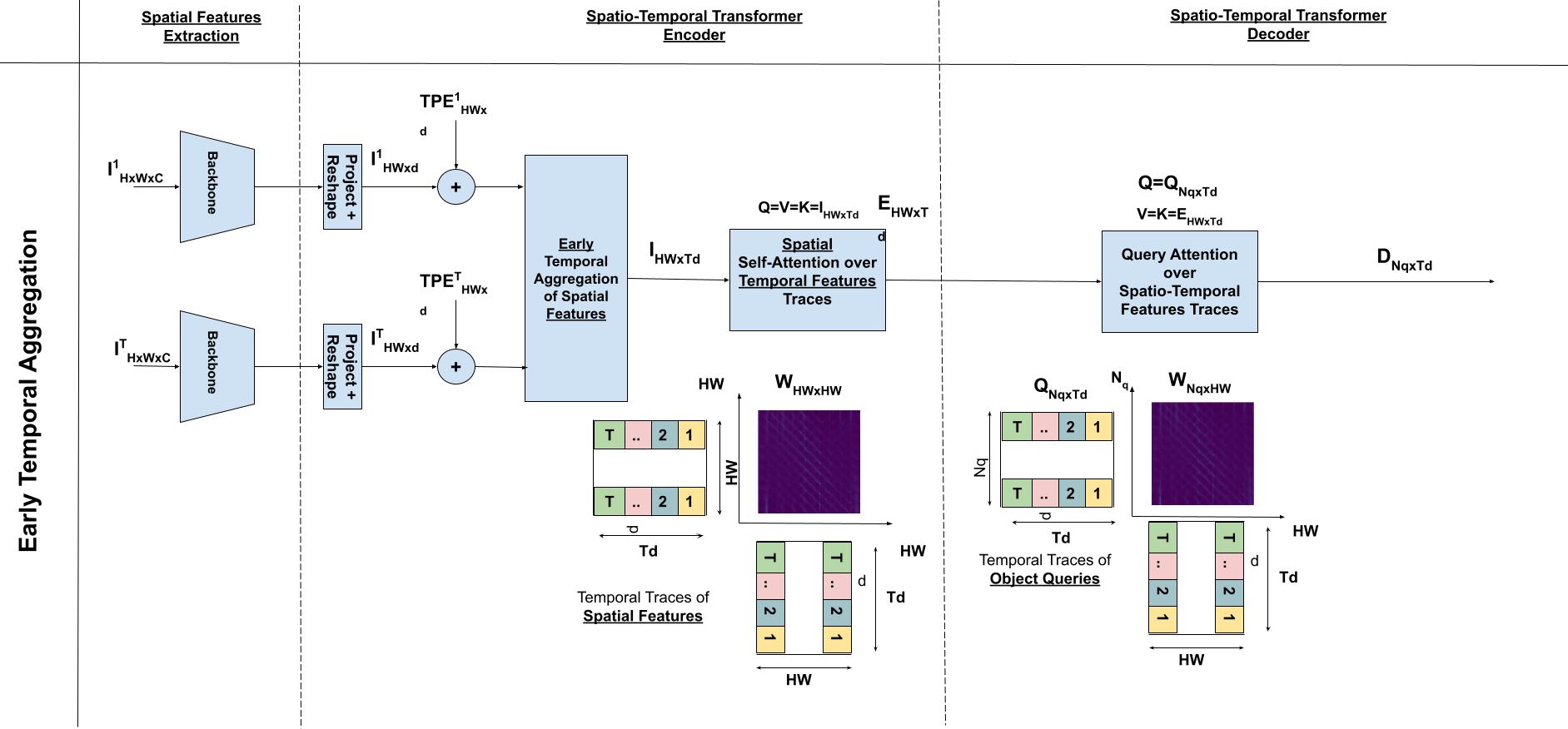}}
\caption{Spatio-Temporal Transformer Model Architecture}
\label{st_modetr}
\end{center}
\end{figure*}

\section{PROPOSED METHOD} \label{sec:methodology}
In this section we present our MTL models. Our approach is based on formulating the MOD and MOSeg problems as a sequence-to-sequence mapping. The input sequence are the consecutive frames, while the output sequence is either the list of objects for MOD, or the segmented frames for MOSeg. We view the input sequence as a Spatio-Temporal sequence, both in space, and time. First, we go through our method of transforming the vanilla encoder-decoder transformer architecture to handle the temporal dimension. Then we present our MTL architecture, where we join the two tasks in one model, evaluating early and late shared transformer layers.

\begin{figure*}
\begin{center}
\centerline{\includegraphics[width=170mm]{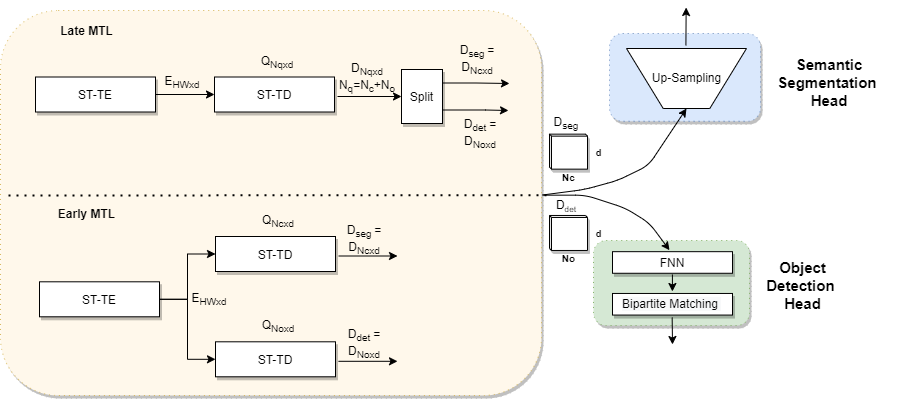}}
\caption{Early and Late architecture options for the shared MTL Transformer}
\label{mtl_early_late}
\end{center}
\end{figure*}
\subsection{Spatio-Temporal Transformer Architecture}
We start by abstracting the spatial 1-step encoder-decoder architecture, referred to here as Vanilla Transformer, like \cite{carion2020end} for object detection and \cite{xie2021segmenting} for segmentation. We view this Vanilla architecture as being formed of the following main steps:

\begin{itemize}
    \item \textbf{Spatial Features Extraction:} using a traditional ConvNet backbone, followed by a $1 \times 1$ convolution to transforms the channels dimension $C=3$ into the hidden dimension $d$. This transforms the input image $I \in \mathbb{R}^{H_1 \times W_1 \times C}$ into  $I \in \mathbb{R}^{H \times W \times d}$, where $H \times W$ represents a coarse spatial grid of spatial features, each represented by a $d$ dimension vector. This feature map is then flattened across the spatial dimensions to be $I \in \mathbb{R}^{HW \times d}$. For simplicity we refer to this as just $I_{HW \times d}$.
    
    \item \textbf{Spatial Transformer:} this maps to the \textbf{Transformer Encoder (TE)} in the DETR paper \cite{carion2020end}. The main objective here is to perform \textbf{Spatial Features Multi-Head Self Attention}. The main idea is treating the spatial features $I_{HW \times d}$ as a sequence of $HW$ spatial features, each of dimension $d$. This can be performed using self-attention mechanism, where we have $Q=V=K=I_{HW \times d}$. First a spatial self-attention map $W \in \mathbb{R}^{HW \times HW}$, or simply $W_{HW \times HW}$ is formed as follows:
    
    $$W_{HW \times HW} = Softmax(QK^T)$$
    
    This is then followed by the transformation of the input $I_{HW \times d}$ using the correlation obtained from the self-attention map as follows:
    
    $$E_{HW \times d} = W_{HW \times HW}I_{HW \times d}$$
    
    \item \textbf{Query Transformer:} this maps to the \textbf{Transformer Decoder (TD)} in the DETR paper \cite{carion2020end}. The main objective here is to map the spatial features into object features based on learn-able \textbf{object queries}: $Q_{o} \in \mathbb{R}^{N_{q} \times d}$, where $N_{q}$ are the number of object queries. In case of segmentation, like \cite{xie2021segmenting}, the object queries will be the same dimension as the number of classes to be segmented $N_c$, while in case of object detection, like \cite{carion2020end}, the object queries will be the number of objects $N_o$. The object queries can be somewhat mapped to the anchors in one-shot ConvNet architectures like YOLO \cite{redmon2017yolo9000}. To obtain the object features, we need to perform Multi-head Attention, with queries are the learn-able object queries $Q=Q_{o}$ and $V=K=E_{HWxd}$ being the spatial features from the encoder. First, the query-spatial features maps are obtained using Multi-head Attention as follows:
    
    $$W_{N_{q} \times HW} = Softmax(Q_{o}K^T)$$
    
    This is then followed by the transformation of the spatial features $E_{HW \times d}$ into object queries features $D \in \mathbb{R}^{N_{q} \times d}$ or simply $D_{N_{q} \times d}$ as follows:
    
    $$D_{N_{q} \times d} = W_{N_{q} \times HW} E_{HW \times d}$$

\end{itemize}

To transform the vanilla 1-step transformer to deal with temporal sequences, then we need to perform the following changes:
\begin{itemize}
    \item \textbf{Spatio-Temporal Features extraction:} we have to first deal with multiple streams of $T$ time steps, each having a spatial feature $I_{HW \times d}$, resulting in $I_{HW \times Td}$ streams. We encode the motion information through the Optical Flow (OF) map highlighting pixels motion. In this approach, we make use of the FlowNet 2.0 \cite{ilg2017flownet} model to compute optical flow. The fusion between appearance (RGB) and motion (OF) is performed on the feature level.

    \item \textbf{Spatio-Temporal Transformer Encoder (ST-TE):} which performs self-attention over the spatial $HW$ dimension, resulting in $E \in \mathbb{R}^{HW \times Td}$.
    
    \item \textbf{Spatio-Temporal Query Transformer Decoder (ST-TD)} which performs the query-to-spatial multi-head attention transformation, resulting in $D \in \mathbb{R}^{N_{q} \times d_{final}}$, where $d_{final}$ is the final dimension after spatio-temporal queries aggregation.
    
\end{itemize}
We adopt the Early Temporal Aggregation, using Multi-Head Self Attention over Spatio-Temporal Features Traces. In this approach, the list of $T$ spatial features $I_{HWxd}$ are aggregated and flattened into $I_{HWxTd}$. This aggregated tensor $I_{HWxTd}$ can thought of as a spatial map of $T$ \textbf{temporal traces of spatial features}, each of dimension $d$, mapped to the spatial locations $H \times W$. This is visualized in Figure \ref{st_modetr}. The ST-TE will then perform multi-head self-attention over the this spatio-temporal map of object features traces. In this case, we have $Q=V=K=I_{HW \times Td}$. The spatio-temporal features traces attention map $W_{HW \times HW} = Softmax(QK^T)$ is then used to obtain the spatio-temporal features $E_{HW \times Td} = W_{HW \times HW}I_{HW \times Td}$.

The ST-TD will perform multi-head query-to-spatio-temporal features traces attention, where $Q \in \mathbb{R}^{N_{q} \times Td}$ and $V=K=E_{HW \times Td}$. The query-spatio-temporal features traces attention map will be $W_{N_{q} \times HW} = Softmax(QK^T)$, resulting in $D_{N_q \times Td} = W_{N_{q} \times HW}E_{HW \times Td}$. This represents the final object queries spatio-temporal features, where $d_{final} = Td$ in this case. For our case, we only process the last frame at $t=T$, since we want to predict the motion in the current frame, given the previos $T$ frames. Thus, we have $D_{N_q \times d}$.

Transformers are originally presented as a replacement to recurrent models, due to their fast parallel encoding nature \cite{vaswani2017attention}. However, this comes at the cost of losing the sequential information of the input. To overcome that, positional encoding embedding was proposed in \cite{vaswani2017attention}. Following on that, the vanilla 1-step DETR \cite{carion2020end} treats the input features as being sequential in the spatial dimension $HW$, which leads to the proposal of Spatial Positional Encoding (SPE). In our spatio-temporal model, a similar encoding is needed to distinguish the temporal sequential information of frames. Hence, we propose a Temporal Positional Encoding (TPE), which is added just before the temporal aggregation takes place, being it early across the spatial features traces $TPE_{HWxd}$ or late across the object queries traces $TPE_{Nqxd}$, see Figure \ref{st_modetr}.

\subsection{Moving object detection output generation head}
The output head simply decodes the object queries features $D_{N_q \times d}$ into the required output. For detection, we will have a number of $N_{q}=N_o$ predicted object bounding box parameters $c_{x}, c_{y}, L, W$ and class.

\subsection{Motion segmentation output generation head}
For segmentation, we will have a mask of dimensions $M_{N_c \times H_{1}W_{1}}$. To transform the decoder output $D_{N_{q} \times d}$ into $M_{N_c \times H_{1}W{1}}$, we undergo reshaping and up sampling as in \cite{xie2021segmenting}. However, this requires to set $d=HW$, so that we can perform the reverse mapping that we did in the ConvNet backbone from the initial image.

\subsection{Multi-Task Learning Transformer Model for Joint Moving Object Detection and Segmentation}
In this section, we discuss our MTL architecture, given the spatio-temporal individual architectures for MOD and MOSeg discussed above. We evaluate two options: 1) early shared transformer encoder, where we have one ST-TE, with output $E_{HW \times Td}$, and two specialized ST-TD, one for each head, and 2) late shared transformer encoder-decoder, with output $D_{N_q \times HW}$ (recall that $d=HW$). Both options are shown in Figure \ref{mtl_early_late} and discussed in the results section.

For the later choice, we have to find a way to split the final decoder output $D_{N_q \times HW}$ for both heads. We propose a joint query structure in the shared transformer decoder, where we set $N_q = N_o + N_c$,  so that the decoded features is a concatenation of both spatio-temporal features for MOSeg and MOD. For output generation, we just split the vector and pass it to the specialized heads.
\begin{figure*}
\begin{center}
\centerline{\includegraphics[width=170mm]{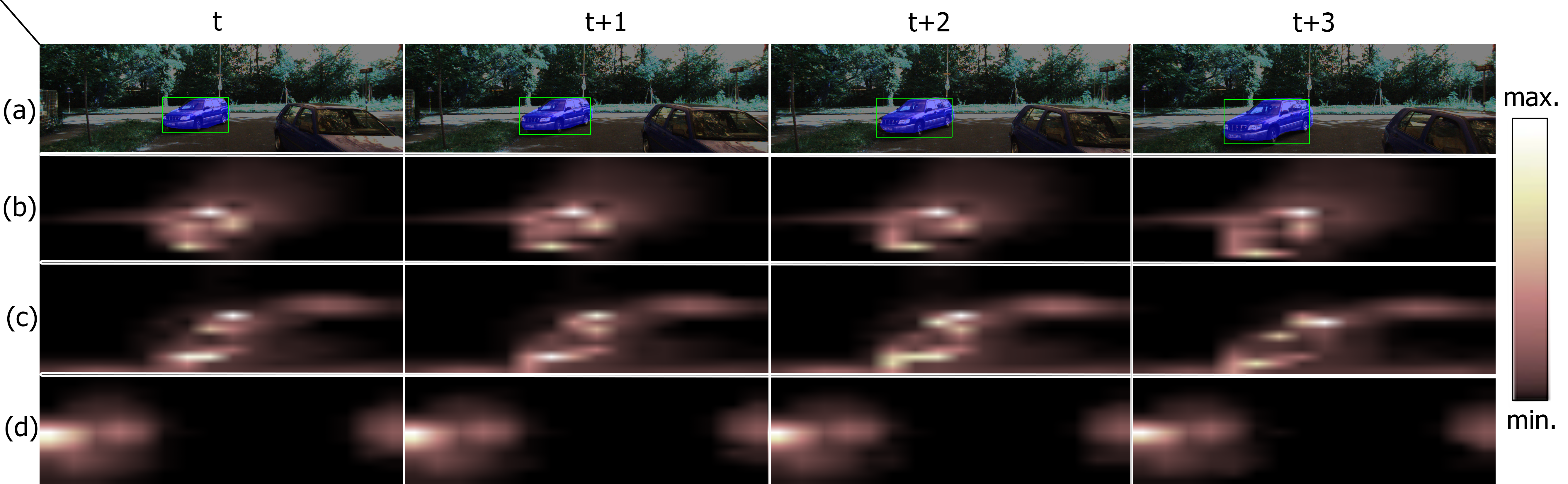}}
\caption{MTL Attention Maps Visualization across different time-stamps. a) shows the MTL output for both detection and segmentation tasks. b) shows the attention maps for object detection task. c) shows the attention maps for moving objects masks. d) shows the attention maps for the background.}
\label{st_modetr:MTL Attention Maps}
\end{center}
\end{figure*}

\section{EXPERIMENTS} \label{sec:experiments}
In this section, we first describe the used datasets. After that, we specify the experimental setup, including all hyper-parameters, and hardware specifications. Finally, We design our experiments to evaluate each of our contributions, in the form of an ablation study to evaluate the impact of each one.

\subsection{Dataset}
There is a huge limitation in publicly available datasets regarding moving object detection. \cite{siam2018modnet} provides 1300 images only with weak annotation for MOD task. \cite{vertens2017smsnet} provides 255 annotated frames only on KITTI dataset, and 3475 annotated frames on Cityscapes dataset \cite{cordts2015cityscapes}.
Thus, We use the extended version \cite{rashed2019fusemodnet} of the publicly available KittiMoSeg dataset \cite{siam2018modnet}. \cite{rashed2019fusemodnet} dataset consists of 12919 frames which are split into 80\% for training, and 20\% for testing. The image resolution is $1242 \times 375$, and the labels determine whether the object is moving or static, includes the object bounding box and the motion mask.

\subsection{Experimental Setup}

We initialize our backbone networks with the weights pre-trained on ImageNet \cite{deng2009imagenet}, then train the whole network for 30 epochs on COCO dataset \cite{lin2014microsoft} while freezing the backbone during the first 10 epochs.
In all our experiments, ResNet-50 \cite{he2016deep} was used as a backbone.
Our network is trained with Adam optimizer \cite{adam} with a scheduled learning rate that is decreased from $1e^{-3}$ to $1e^{-5}$, the whole network is end-to-end trained with learning rate exponentially decayed.
We train a total of 200 epochs, using a warm-up learning rate of $1e^{-3}$ to $5e^{-3}$ in the first 5 epochs, and a learning rate exponentially decayed from $1e^{-3}$ to $1e^{-5}$ in the rest of epochs.
$512 \times 512$ resolution images have been used across all the experiments and Td, that represent number of temporal frames that we are using, is set to two.
Our approach is implemented in Python using PyTorch framework on two PCs with Intel Xeon(R) 4108 1.8GHz CPU, 64G RAM, Nvidia Titan-XP.

\subsection{Results} 
Our evaluation first starts with comparing both options of the MTL architecture: Early vs. Late. Then we use the best option to evaluate the value of MTL versus the individual heads; MOD and MOSeg.

\subsection{Early vs. Late MTL evaluation}
In this experiment we evaluate the early vs. late architectures. In the early architecture, the shared transformer encoder is followed by specialized decoders and heads. For each decoder we have $N_o = 100$, and $N_c = 2$.

Results in Table \ref{tab:ST-MTL-early-late} show clear advantage for the late architecture. While the early MTL provide an advantage of around 0.7\% IoU over the individual ST-MOSeg model, it falls behind the individual ST-MOD by around 4 \% mAP (see Table \ref{tab:ST-DET-SEG-MTL}. On the other hand, the late MTL model is better than the individual models on both tasks by 2 \% IoU and 1.5 \% mAP in MOSeg and MOD respectively.

\begin{table}[t]

\begin{center}
\begin{tabular}{l|c|c|c|c}
\hline
Method & $mAP_{Total}$ & $AP_{50}$ & $AP_{75}$ & $IoU$ \\
\hline
MTL-Early          & 36.8\% & 60.7\% & 40.6\% & 79.6\%   \\
MTL-Late           & \textbf{42.8\%}  & \textbf{64.9\%} & \textbf{50.1\%} & \textbf{80.9\%}       \\
\hline
\end{tabular}
\end{center}
\caption{Comparing the Early variant of our MTL architecture, and the Late variant of our MTL architecture.}
\label{tab:ST-MTL-early-late}
\end{table}

While the early encoder sharing enables the common spatial features to be shared across both tasks, the later architecture further enables query features to be shared as well. For our tasks, semantic masks queries and object queries naturally have some correlations. Same as in Panoptic segmentation in DETR \cite{carion2020end}, the object queries could be used to produce segmented masks, in our case, joint queries can be used to encode object and segmentation masks. Thanks to our joint task query method with $N_q = N_o + N_c$, we are able to split the decoder vector to the specific output heads. 

\subsection{MTL vs. individual models evaluation}
To show the value of our MTL architecture, we compare it against the individual tasks models. We follow the late MTL architecture due its superior performance. In all the three architectures we have an early spatio-temporal aggregation. We refer to the segmentation model as ST-MOSeg, the detection as ST-MOD and the joint as ST-MTL in Table \ref{tab:ST-DET-SEG-MTL}.

\begin{table}[t]

\begin{center}
\begin{tabular}{l|c|c|c|c}
\hline
Method & $mAP_{Total}$ & $AP_{50}$ & $AP_{75}$ & $IoU$ \\
\hline
ST-MOD              & 41.3\% & \textbf{65.5\%} & 47.1\% & N/A                          \\
ST-MOSeg           & N/A  & N/A & N/A & 78.9\%       \\
ST-MTL            & \textbf{42.8\%}    & 64.9\%    & \textbf{50.1\%}   & \textbf{80.9\%}         \\
\hline
\end{tabular}
\end{center}
\caption{Comparing the Detection architecture, Segmentation architecture, and MTL architecture.}
\label{tab:ST-DET-SEG-MTL}
\end{table}

Results in Table \ref{tab:ST-DET-SEG-MTL} show that the late MTL model is better than the individual models on both tasks by 2 \% IoU and 1.5 \% mAP in MOSeg and MOD respectively. This result proves the positive effect of joint learning of both tasks, over the individual models. The shared transformer encoder-decoder later architecture maximizes the shared learning, which leads to the improvement of both heads.

\subsection{Bench-marking against state-of-the-art}

\begin{table}[t]
\begin{center}
\begin{tabular}{l|c}
\hline
Method & $IoU$ \\
\hline
RST-MODNet-LSTM-Late              & 69.5\%  \\
RST-MODNet-LSTM-Multistage        & 71.4\%       \\
ST-MTL    (ours)                       & \textbf{80.9\%}   \\
\hline
\end{tabular}
\end{center}
\caption{Quantitative results on KITTI  MOD \cite{siam2018modnet} dataset in terms of intersection over union (IoU) for the segmentation task compared to state-of-the-art methods.}
\label{tab:Bench-marking against state-of-the-art}
\end{table}

Table \ref{tab:Bench-marking against state-of-the-art} shows a comparison between our approach and state-of-the-art baseline methods.
RST-MODNet \cite{ramzy2019rst} has two models, the RST-MODNet-LSTM-Late uses ConvLSTM at the decision level before softmax layer where the network learns to use time information before the final classification is done while the RST-MODNet-LSTM-Multistage uses several ConvLSTM layers across the network at three different stages.
For fair comparison we adapted the original RST-MODNet where four time-steps were used in their experiments while we are taking only two frames into consideration, therefore we set Td to two in all our experiments.
As shown at table \ref{tab:Bench-marking against state-of-the-art} our novel approach is better than the state-of-art methods by 9.5\%.

\subsection{Visualizing MTL attention maps}

Figure \ref{st_modetr:MTL Attention Maps} demonstrates the MTL attention maps visualization across different time-stamps. where row (a) shows the MTL output for both detection and segmentation tasks, row (b) shows the attention maps for the most confident query that is responsible for detecting the moving object, row (c) shows the attention maps for semantic segmentation task where there is an attention map for each class, in our experiment there is only one class of interest that is the moving vehicles and row (d) shows the attention maps for the background, accordingly there are two attention maps responsible for outputting the segmentation mask.
\section{CONCLUSIONS} \label{sec:conc}
In this paper we presented a Spatio-Temporal Multi-Task Learning Transformer based architecture for joint MOD and MOSeg. The Spatio-Temporal MOSeg Transformer architecture is a novel one, showing state-of-the art performance. We compared the MTL setup against the individual tasks, which shows 1.5\% mAP improvement on MOD and 2\% IoU on MOSeg. Moreover, our results suggests a clear advantage for our proposed late joint tasks query transformer decoder, over the early shared encoder, specially on the detection head. There is also another advantage on the dimension of fast inference in the shared MTL architecture over the individual models, which will almost doubles the inference time and the memory footprint over the shared model. The saving in time and memory is maximized in the late MTL architecture, which is possible thanks to our proposed joint tasks queries decoder.





\bibliographystyle{ieee}
\bibliography{references/egbib}

\end{document}